\documentclass[letterpaper, 10 pt, conference]{ieeeconf}  

\IEEEoverridecommandlockouts                              

\usepackage{graphics} 
\usepackage{graphicx}
\usepackage{subcaption}
\usepackage[hyphens]{url}
\usepackage{amsmath} 

\title{\LARGE \bf
Input-gated Bilateral Teleoperation: An Easy-to-implement\\
Force Feedback Teleoperation Method for Low-cost Hardware
}

\author{Yoshiki Kanai$^{1}$, Akira Kanazawa$^{1}$, Hideyuki Ichiwara$^{1}$, \\ 
Hiroshi Ito$^{1}$, Naoaki Noguchi$^{1}$ and Tetsuya Ogata$^{2}$
\thanks{$^{1}$Yoshiki Kanai, Akira Kanazawa, Hideyuki Ichiwara, Hiroshi Ito, and Naoaki Noguchi are with Research \& Development Group, Hitachi, Ltd., Ibaraki, 312-0034,
Japan.
        {\tt\small yoshiki.kanai.fx@hitachi.com}}%
\thanks{$^{2}$ Tetsuya Ogata is with Department of Intermedia Art and Science School of Fundamental Science and Engineering, Waseda University, Tokyo, 169-855, Japan.
        {\tt\small ogata@waseda.jp}}%
}

\begin{document}

\maketitle
\thispagestyle{empty}
\pagestyle{empty}

\begin{abstract}
Effective data collection in contact-rich manipulation requires force feedback during teleoperation, as accurate perception of contact is crucial for stable control. 
However, such technology remains uncommon, largely because bilateral teleoperation systems are complex and difficult to implement. 
To overcome this, we propose a bilateral teleoperation method that relies only on a simple feedback controller and does not require force sensors. 
The approach is designed for leader-follower setups using low-cost hardware, making it broadly applicable. Through numerical simulations and real-world experiments, 
we demonstrate that the method requires minimal parameter tuning, yet achieves both high operability and contact stability, outperforming conventional approaches. 
Furthermore, we show its high robustness: even at low communication cycle rates between leader and follower, 
control performance degradation is minimal compared to high-speed operation. 
We also prove our method can be implemented on two types of commercially available low-cost hardware with zero parameter adjustments. 
This highlights its high ease of implementation and versatility. We expect this method will expand the use of force feedback teleoperation systems on low- cost hardware. 
This will contribute to advancing contact-rich task autonomy in imitation learning.
\end{abstract}

\section{Introduction}
With advancements in deep learning and its applications to robotics, 
robots are now capable of autonomously performing complex tasks 
that were previously considered difficult to achieve. 
In particular, imitation learning \cite{Hussein17,Bahl22}, a method that converts human demonstrations into data 
to learn behavioral policies, enables adaptive robot control. 
The introduction of low-cost and accessible open-source hardware \cite{ALOHA, GELLO} has lowered both economic 
and technical barriers in imitation learning, garnering attention for accelerating research on robotic foundation models \cite{RTX,pi0}. 
For data collection in imitation learning, leader-follower teleoperation systems \cite{Kim20} are widely utilized, 
where the follower robot is controlled to track the state of the leader robot operated by the user, enabling intuitive robot control. 
Furthermore, bilateral teleoperation \cite{yokokoji94, force-reflection}, which dynamically couples the mechanical systems of the leader and follower, 
allows for the implementation of teleoperation systems with force-feedback. By leveraging the force information obtained from such systems, 
policy learning methods have been developed to effectively utilize this data, expanding the applicability of robots to contact-rich tasks  \cite{Saigusa22, Bi-ACT, FACTR, tsuji25}.

However, teleoperation systems with force-feedback have not yet achieved widespread adoption. 
One of the reasons for this is the difficulty of implementing force-feedback teleoperation methods for low-cost hardware. 
For example, the bilateral teleoperation method used in reference \cite{FACTR} assumes that joint torque sensors are installed on the follower robot. 
Currently, few robotic products or low-cost hardware come equipped with torque sensors as standard, resulting in limited generality. 
The acceleration-based bilateral controller (ABC) \cite{iida04, ABC} used in references \cite{Saigusa22, Bi-ACT} does not require force sensors 
and has demonstrated excellent control performance through high controllability achieved by acceleration control and high responsiveness enabled 
by broadband reaction force estimation \cite{Sariyildiz15}. 
However, ABC requires careful parameter tuning due to its reliance on the design of filters, identification of physical models, 
and control parameters for both position and force controllers.

Feedback controllers in prior studies are implemented on high-level systems built on general-purpose PCs. 
For enhanced performance, low-level systems like microcontrollers or FPGAs are generally desired, 
offering high cycle rates and real-time capabilities \cite{Franklinet98}. 
However, low-cost hardware actuators typically have limited low-level functionalities. 
This necessitates implementing complex controllers from prior research on high-level systems. 
Reported feedback controller cycle rates (500-1000 Hz) \cite{Saigusa22, Bi-ACT, FACTR} are insufficient due to high-level 
to low-level communication limitations. 
Consequently, control parameters require trial-and-error adjustment to achieve maximum performance 
without destabilizing the system.
Although real-time kernels or dedicated control boards can improve high-level system cycle rates, 
these approaches increase development effort and cost, undermining low-cost hardware affordability. 
Furthermore, cycle rate drops caused by PC specifications or actuator count lead to non-transferable parameters and low robustness. 
While some bilateral teleoperation methods can be implemented using simple low-level feedback controllers, 
they often provide insufficient control performance in operability and contact stability \cite{yokokoji94, force-reflection}. 
Thus, implementing high-performance bilateral teleoperation remains challenging, requiring specialized motion control knowledge 
and maintaining a high barrier to entry.

In this paper, we propose a bilateral teleoperation method that is easily implementable for low-cost hardware. 
This method is referred to as Input-Gated Bilateral Teleoperation (IGBT). 
The motivation for introducing IGBT is to enable easier implementation of bilateral teleoperation for low-cost hardware, 
thereby promoting the adoption of teleoperation systems with force-feedback. 
IGBT does not require force sensors and can be implemented using a simple low-level feedback controller independent of high-level system operating environments, 
making it highly implementable. The effectiveness of IGBT is validated through numerical simulations, experiments with actual robots, 
and tests using commercially available low-cost hardware. As a result, IGBT achieves a balance between high operability and contact stability, 
outperforming two other bilateral teleoperation methods with comparable parameter tuning efforts. 
Additionally, IGBT maintains control performance even when the cycle rate of the high-level system is slow. Furthermore, 
by utilizing the default parameters of position controllers in commercially available actuator products, 
IGBT enables teleoperation systems with force-feedback to be implemented with effectively zero parameter tuning effort.

The contributions of this paper are as follows:
\begin{itemize}

\item Proposal of a bilateral teleoperation method for low-cost hardware that does not require force sensors and can be easily implemented using only low-level controllers.
\item Verification that the proposed method demonstrates high operability and contact stability, even when the cycle rate of the high-level system is not sufficiently fast.
\item Validation of the applicability of proposed method to commercially available low-cost hardware.

\end{itemize}

\section{Related Works}

\subsection{Data collection approaches for imitation learning}
In imitation learning, three types of data collection systems have been proposed: 
(1) leader-follower type \cite{mobileALOHA, sasagawa20}, 
(2) handheld type \cite{UMI, ForceMimic}, and 
(3) exoskeleton type \cite{Exo, ACE}. 
In leader-follower type, two robots are utilized, where the follower robot mimics the movements of 
the leader robot controlled by the operator to provide demonstrations. 
A key advantage of this approach is that if the leader and follower robots share the same configuration, 
motion synchronization can be easily achieved through simple mapping in the joint space.
On the other hand, this approach requires the operator to demonstrate interactions with the environment through the leader robot, 
making force-feedback indispensable, particularly for contact-rich tasks. 
If force-feedback is not adequately transmitted, it becomes difficult to consistently reproduce the forces applied to the environment in the collected data, 
which may ultimately degrade the learning performance of imitation learning. 
Therefore, to enable high-quality data collection for contact-rich tasks, it is crucial to implement force-feedback teleoperation through hardware innovations, 
such as the integration of force sensors, or control strategies such as bilateral control.

Handheld type and exoskeleton type are proposed as alternatives that allow the operator 
to more directly experience interaction with the environment during data collection. 
In handheld type, operators can directly perceive interactions with the environment through the device, 
enabling natural and high-precision data collection. 
However, the collected data in the end-effector space must be converted into the joint space, 
and structural constraints of the robot, such as its range of motion and degrees-of-freedom, 
may lead to actions that the robot cannot physically perform.
Exoskeleton type, similar to leader-follower type, 
enable data collection while allowing the operator to directly interact with the environment 
by wearing the exoskeleton, facilitating the acquisition of more natural data.
However, due to the nature of being worn on the human body, 
exoskeleton type require careful design to ensure appropriate size, weight, 
and unobstructed human movement, which imposes limitations on the applicable robotic systems.

In this paper, we address these challenges by providing a method that facilitates the easy implementation of force-feedback in leader-follower systems, 
thereby promoting high-quality data collection for contact-rich tasks.

\subsection{Actuator modules for low-cost hardware}
We focus on actuator modules commonly used in low-cost hardware platforms. 
For example, robotic systems such as ALOHA \cite{ALOHA} and GELLO \cite{GELLO} employ Dynamixel modules 
manufactured by ROBOTIS \cite{dynamixel}. In recent years, actuators known as Quasi-Direct Drive (QDD) \cite{wensing17, blue, perez24}, 
which typically feature low gear ratios around 1:10, have also garnered significant attention.
These actuator modules have two notable features: (1) high backdrivability and (2) ease of use.
Regarding backdrivability, these actuators are equipped with natural compliance, 
allowing them to withstand high-impact and high-frequency contact forces. 
Additionally, due to their low friction, the relationship between current and torque is easily modeled, 
making them suitable for contact-rich tasks.
Regarding ease of use, these actuators have a compact design that integrates the drive system 
and control system, enhancing the scalability of robot design and promoting plug-and-play development. 
These modules come pre-implemented with basic motion controllers, 
such as position control and current control, enabling users to easily control robots 
without the need to design low-level controllers.

On the other hand, these actuators often lack support for local communication between modules or direct connections with external sensors. 
Consequently, implementing force feedback control or bilateral teleoperation necessitates collecting data from each module in a high-level system, 
calculating control commands, and then transmitting them to the low-level system. When a feedback controller resides in the high-level system, 
communication with the low-level system can hinder achieving high cycle rates, thereby limiting the improvement of control performance.
Given these actuator characteristics, this study proposes an approach that avoids placing the feedback controller in the high-level system. 
Instead, we present a method for realizing bilateral teleoperation solely by utilizing functionalities implemented within the actuator module's low-level controller.

\subsection{Bilateral Teleoperation}
Bilateral teleoperation is a control method in which the state variables (e.g., position and force) 
of the leader and the follower system are synchronized, enabling unified control of both systems. 
This allows not only position tracking but also the reproduction of action-reaction dynamics of forces. 
Several methods have been proposed, with the following four being the most representative: 
(1) the symmetric position servo type bilateral teleoperation (SPBT),
(2) the force reflection type bilateral teleoperation (FRBT), 
(3) the force reflecting servo type (FSBT) \cite{yokokoji94, force-reflection}, and 
(4) Acceleration-based bilateral control (ABC) \cite{iida04, ABC}.

The SPBT involves both the leader and follower exchanging their position information and performing position control. 
This approach does not require additional sensors, making it simple to implement and highly stable during contact. 
However, since position controllers are typically designed with high servo stiffness, this method tends to behave rigidly in response to user input forces, 
resulting in a stiff operational feel. Lowering the control gains to improve operability can reduce the user ability to sense reaction forces during environmental contact, 
leading to a trade-off between operability and contact sensitivity.

The FRBT transmits force information obtained from the position-controlled follower directly as a command to the leader. 
Since the leader lacks a position feedback loop, this approach achieves high operability. 
A simple implementation involves obtaining motor current values from the follower and inputting them as control commands to the current-controlled leader. 
This method can be realized using only the built-in current sensors, which is a significant advantage. 
However, relying solely on current control results in low disturbance rejection performance, making contact stability highly susceptible to communication delays and noise.

The FSBT equips both the leader and follower with force sensors and constructs a force feedback loop on the leader side. 
Similar to the FRBT, it offers high operability, but by forming a closed-loop outside the current control, it achieves higher contact stability. 
However, this approach requires force sensors and is sensitive to noise and response characteristics, making gain tuning hard.

The ABC targets the control of four variables: the positions and forces of both the leader and the follower. 
Using acceleration control with a disturbance observer (DOB) \cite{DOB}, this method achieves high controllability 
and high responsiveness through broadband reaction force estimation by reaction force observer (RFOB) \cite{RFOB}. 
However, it requires the tuning of numerous design parameters, including model identification, observer gains, and position and force control gains, 
resulting in significant tuning effort. 
Moreover, due to the influence of communication delays and noise, suitable parameter values become environment-dependent, leading to potential fluctuations in performance.

A fundamental trade-off exists between control performance and ease of implementation in bilateral teleoperation methods. 
For instance, while methods like SPBT and FRBT are easy to implement, their performance is limited. Conversely, 
methods such as FSBT and ABC offer high performance but are complex to implement.
Even with sophisticated control logic, inherent hardware limitations in low-cost systems can cap achievable performance, 
creating a bottleneck. These include, for example, actuator responsiveness, sensor precision, and low-level processor power. 
In such constrained environments, over-pursuing performance often complicates implementation without yielding commensurate 
gains for the invested effort.
Therefore, this study prioritizes securing practically sufficient performance without compromising ease of implementation, 
rather than absolute performance maximization.

\begin{figure}[t]
        \centering
        \includegraphics[width=0.7\columnwidth]{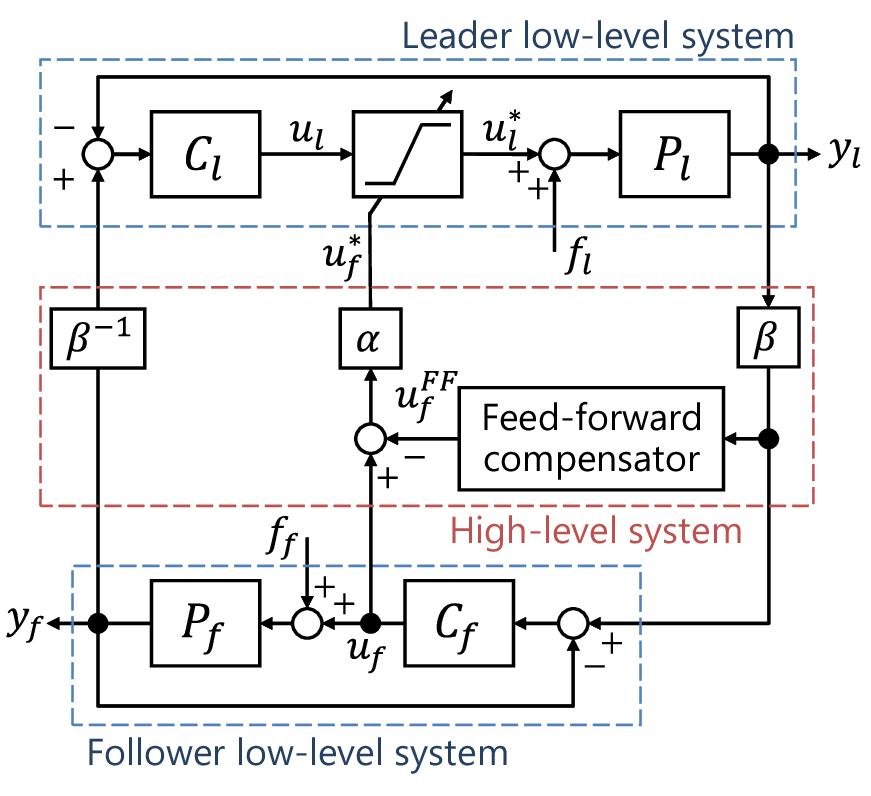}
        \vspace*{-4mm}
        \caption{Block diagram of the generalized IGBT. Both the leader and follower are operated under position control, and the leader control input $u_l$ is gated based on the follower control input $u_f$. The high-level system does not have a feedback controller.}
        \vspace*{-4mm}
        \label{fig:block}
\end{figure}

\section{Proposed Method}
Fig. \ref{fig:block} shows the block diagram of the generalized IGBT system, which can be extended to a micro-macro bilateral control system \cite{micromacro}. 
Both the leader system and the follower system construct a position control loop, 
and a limiter on the control input is applied only to the leader system. 
Subscript $l$ denotes variables of the leader system, while subscript $f$ denotes those of the follower system. 
Here, $y$ represents the system position, $C$ the position controller, $P$ the plant, $f$ the external force applied to the system, 
$u$ the control input, and $u^*_l$ the control input after limitation. 
$\alpha$ is scaling factor for force and $\beta$ is for position.  By setting these to arbitrary values, bilateral teleoperation between a leader and a follower with different dimensions or output capabilities can be achieved.
$u_f^{FF}$ is a variable used for feed-forward compensation of disturbances, 
such as kinetic friction torque or gravitational torque of the robot arm, that can be modeled in advance.
In this case, $y$ and $u$ are expressed by the following equations.
\begin{align}
y_l&=P_l(f_l+u_l^*) \\
y_f&=P_f(f_f+u_f) \\
u_l&=C_l(\beta^{-1} y_f-y_l)\\
u_f&=C_f(\beta y_l-y_f)
\end{align}
The upper and lower bounds of the control input limiter are set to the absolute value of $u_f^*$, as defined by the following equation.

\begin{align}
u_f^*&=\alpha(u_f-u_f^{FF}) \\
u_l^*&=
\begin{cases}
        -|u_f^*|, & u_l<-|u_f^*| \\
        u_l,    & -|u_f^*| \leq u_l \leq |u_f^*| \\
        |u_f^*|, & |u_f^*|<u_l
\end{cases}
\label{eq:limit}
\end{align}
With the above configuration, the only control parameters that require tuning in the proposed method are the position controllers $C_l$ and $C_f$, which is equivalent to the SPBT. 
Furthermore, if the leader low-level system provides a function for limiting control inputs (e.g., current, torque, or PWM), 
the high-level system only transmits and receives the state variables of each system and calculates the feedforward compensator, 
without incorporating a feedback controller.
As shown in Eq. \ref{eq:limit}, since the method gates the control input, it is referred to in this paper as ``Input-gated'' Bilateral Teleoperation (IGBT).

\begin{figure}[t]
        \centering
        \includegraphics[width=0.8\columnwidth]{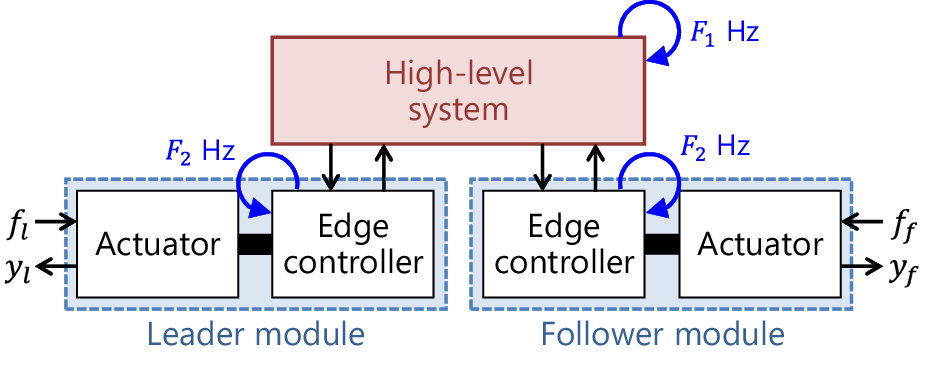}
        \vspace*{-4mm}
        \caption{Configuration diagram of bilateral teleoperation using an actuator module for low-cost hardware.}
        \vspace*{-4mm}
        \label{fig:system_struct}
\end{figure}

The behavior of the IGBT is explained below. For simplicity, let $\alpha=1$, $\beta=1$, and $u_f^{FF}=0$.
The IGBT has been configured with modifications to improve the operability of the SPBT. 
Fig. \ref{fig:system_struct} shows the implementation diagram of bilateral teleoperation using actuator modules for low-cost hardware. 
The edge controller within the actuator module operates at a cycle rate of $F_2$ [Hz], 
while the high-level system operates at a cycle rate of $F_1$ [Hz], with the condition that $F_1<F_2$.
When the SPBT is implemented, the edge controllers are position controllers. 
When an external force $f_l$ is applied to the leader system while the follower system is not in contact with the environment (i.e., $f_f=0$, hereafter referred to as free motion), 
the position displacement of the leader system is transmitted to the follower system via the high-level system. 
Before the position displacement of the follower system returns to the leader system, 
the leader edge controller generates a control input $u_l$ to cancel the external force $f_l$, 
which degrades operability. Here, if the control input $u_l$ during free motion can be reduced, it is expected to improve operability. 
On the other hand, when the follower system is in contact with the environment (i.e., $|f_f>0|$, hereafter referred to as contact motion), 
it is sufficient for the leader system to follow the position displacement of the follower system, 
and if the position controller functions, contact stability will be high. 
That is, in the SPBT, by limiting the control input $u_l$ during only free motion, 
both high operability and high contact stability can be achieved. In the IGBT, 
the difference between the two motions is focused on the control input $u_f$ of the follower system. 
Since the follower edge controller is also a position controller, 
the control input $u_f$ changes dynamically in response to variations in the external force $f_f$. 
Therefore, by dynamically limiting $u_l$ based on $u_f$ as shown in Eq. \ref{eq:limit}, 
the limiting amount will automatically change in response to the motion, 
enabling both high operability and high contact stability in the IGBT.

Alternatively, estimating free/contact states to switch controllers or dynamically adjusting the leader's position 
controller gain is possible. However, these methods require significant effort for state estimation or gain adjustment, 
hindering ease of implementation. FRBT, with comparable adjustment effort, uses a current controller for the leader, 
potentially compromising motion stability. Indeed, reference \cite{FACTR} introduces a damping term to suppress oscillations, 
requiring additional coefficient adjustment.
In the IGBT, the outer-loop position controller within the low-level system ensures the overall system's motion stability. 
Notably, even when the high-level system's cycle rate fluctuates or decreases due to factors like PC specifications or 
the number of connected actuators, the low-level system's operating cycle remains unaffected. This significantly contributes 
to the system's overall stability. Consequently, low-level system control parameters can be directly applied across 
varying high-level system operating environments, ensuring high robustness.

In this paper, we propose an IBGT that achieves both high operability and high contact stability merely by adjusting the position controllers of the leader and follower. 
This is accomplished by dynamically varying the limitation amount of the leader's control input $u_l$. 
The effectiveness of the proposed IBGT will be demonstrated through numerical simulations and real-world experiments in the next chapter.

\section{Evaluations}
\subsection{Numerical simulation}
First, the effectiveness of the IGBT is verified through numerical simulation. 
The system configuration follows Fig. \ref{fig:system_struct}, which assumes the use of an actuator designed for low-cost hardware. 
The response of the proposed method is compared with that of the SPBT and the FRBT, 
both of which require a similar level of parameter tuning effort. 
Here, $F_1<F_2$, and the command values sent to the edge controllers are assumed to be held using zero-order hold. 
For simplicity, we assume that the controlled system consists of two identical electric motors, i.e.,
$P_l=P_f=P$, $C_l=C_f=C$, $\alpha=1$, $\beta=1$, and $u_f^{FF}=0$, 
and that current control functions ideally, such that current and torque are in an ideal proportional relationship. 
The plant $P$ and controller $C$ are defined by the following equations.
\begin{align}
        P&=\frac{1}{Js^2+Ds}\\
        C&=K_ds+K_p
\end{align}
Here, $J$ represents the moment of inertia, and $D$ is the viscous damping coefficient. 
The controller $C$ is a PD controller, with two parameters: the proportional gain $K_p$ and the derivative gain $K_d$.

The conditions for the numerical simulation are described as follows. First, at simulation time $t=0.0$ s, 
a step external force of $f_l=1.0\times10^{-2}$ N$\cdot$m is applied to the leader system. Next, 
it is assumed that the follower system collides with the environment at time $t=1.0$ s. 
That is, the position of the follower system at $t=0.0$ s is set as the environmental position $y_\mathrm{env}=y_f (1.0)$, 
and the external force $f_f$ acting on the follower system is defined by the following equation.
\begin{align}
        e(t)=
        \begin{cases}
            0, & t<1 \\
            y_\mathrm{env}-y_f(t), & t \geq 1
        \end{cases}\\
        f_f(t)=
        \begin{cases}
            K_\mathrm{env} \, e(t), & e(t)<0 \\
            0, & e(t) \geq 0
        \end{cases}
\end{align}
Here, $K_\mathrm{env}$ represents the environmental stiffness.
The various parameters are summarized in the Tab. \ref{tab:param}, 
and in order to examine the effect of differences in the high-level system cycle rate, 
simulations were conducted with  set to 100 Hz and 1 kHz.
Under these conditions, the time response of the follower system position $y_f$ when an external force is applied is analyzed for each of the following methods: 
SPBT ($u_l^*=u_l$), FRBT ($u_l^*=-u_f$), and IGBT (Eq. \ref{eq:limit}).
\begin{table}[t]
        \centering
        \caption{Simulation parameters.}
        \vspace*{-1mm}
        \begin{tabular}{ccc}
            \hline
            Parameter & Contents & Value\\
            \hline
            $J$ & Moment of inertia & 1.0$\times$10$^{-3}$ [kg$\cdot$m$^2$] \\ %
            $D$ & Viscous friction constant & 1.0$\times$10$^{-2}$ [kg$\cdot$m$^2\cdot$s/rad] \\ %
            $K_p$ & P gain for controller & 1.0$\times$10$^{1}$\\
            $K_d$ & D gain for controller & 2.0\\
            $K_\mathrm{env}$ & Stiffness of Environment & 1.0$\times$10$^4$ [N$\cdot$m/rad]\\
            $F_2$ & Cycle rate for edge controller & 1.0$\times$10$^4$ [Hz] \\
            \hline
        \end{tabular}
        \label{tab:param}
        \vspace*{-5mm}
\end{table}

\begin{figure}[t]
    \centering
    \begin{minipage}{0.492\columnwidth}
        \centering
        \includegraphics[width=\columnwidth]{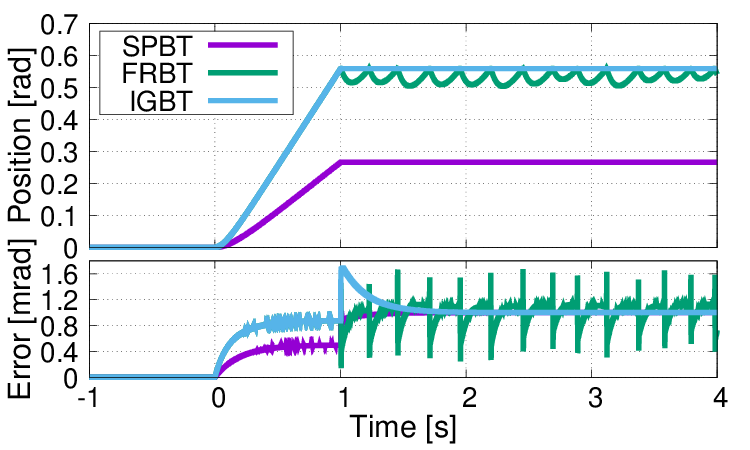}
        \subcaption{}
        \label{fig:sim1000}
    \end{minipage}
    \begin{minipage}{0.492\columnwidth}
        \centering
        \includegraphics[width=\columnwidth]{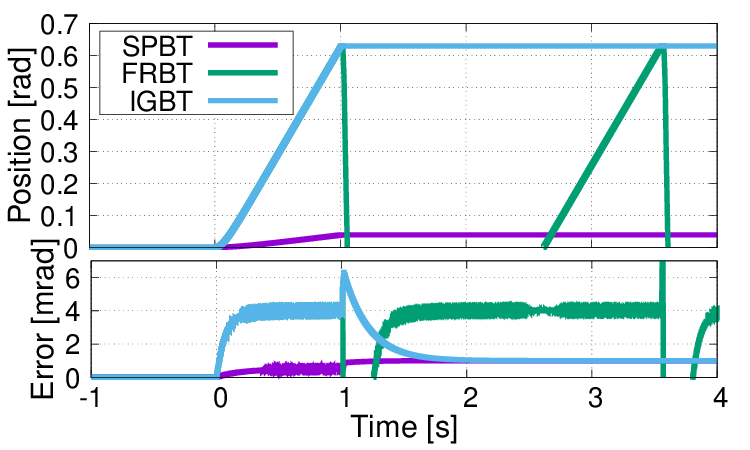}
        \subcaption{}
        \label{fig:sim100}
    \end{minipage}
    \caption{Numerical simulation results. The time responses of the follower system position $y_f$ and position error $y_l-y_f$ are plotted for the three methods: SPBT, FRBT, and IGBT. (a) $F_1=1$ kHz. (b) $F_1=100$ Hz.}
    \label{fig:sim_result}
\end{figure}

Fig. \ref{fig:sim_result} shows a comparison of the time responses of follower position $y_f$ 
and position error between leader and follower $y_l-y_f$ obtained from the numerical simulations.
In the free motion phase where $0 \leq t \leq 1$, the fact that only SPBT exhibits a gentle slope indicates 
that the system is less responsive to the same external force, resulting in poor operability. 
Since FRBT and IGBT show consistent responses, the operability of IGBT is equivalent to that of FRBT.
In the contact motion phase where $t>1$, FRBT exhibits a bouncing behavior, indicating low contact stability. 
On the other hand, SPBT and IGBT remain stationary, and the positional error between the leader and follower 
converges to 1 mrad in both cases, demonstrating that the contact stability of IGBT is equivalent to that of SPBT.

Next, we compare the responses of the high-level systems under different cycle rates. 
In the free motion phase where $0 \leq t \leq 1$, SPBT shows an 84\% decrease in speed, 
indicating a significant reduction in operability. On the other hand, 
FRBT and IGBT exhibit approximately a 12\% increase in speed, showing a more gradual change in operability compared to SPBT.
Subsequently, in the contact motion phase where $t>1$, FRBT demonstrates an increase in the rebound amount after collision, 
resulting in a significant decline in contact stability. Conversely, both SPBT and IGBT converge to a positional error of 1 mrad, 
confirming that they maintain high contact stability.

Based on the above results, IGBT has been shown to achieve operability equivalent to FRBT and contact stability equivalent to SPBT, 
with the same level of parameter adjustment effort as SPBT. Additionally, it has been demonstrated that the operability 
and contact stability of IGBT are robust against changes in the cycle of the high-level system.

\subsection{Experiment on actuator modules}
\begin{figure}[t]
        \centering
        \includegraphics[width=0.7\columnwidth]{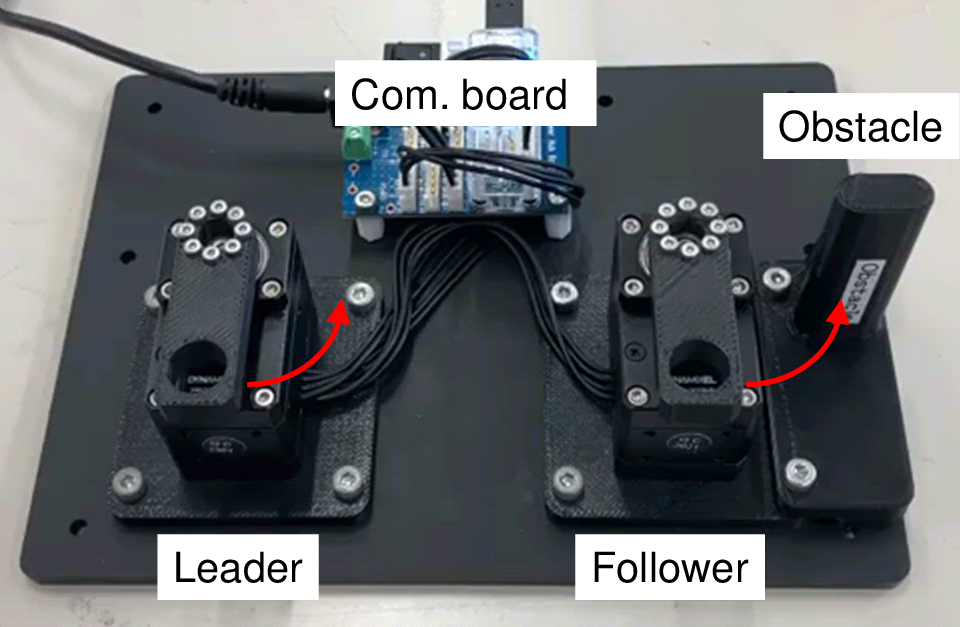}
        \caption{experimental environment. A pair of Dynamixel XM430-W350-R from ROBOTIS are used.}
        \vspace*{-4mm}
        \label{fig:exp_env}
\end{figure}

The effectiveness of IGBT is validated by implementing it on low-cost hardware. 
First, bilateral teleoperation is evaluated between a pair of identical actuators. 
The experimental setup is shown in the Fig. \ref{fig:exp_env}. The actuator modules used are Dynamixel XM430-W350-R from ROBOTIS. 
To facilitate the application of external forces, links are attached to the output shafts of each actuator. 
Additionally, an obstacle is placed within the movable range of the follower link to simulate contact with the environment.

\begin{figure*}[t!]
    \centering
    \begin{minipage}{0.245\textwidth}
        \centering
        \includegraphics[width=\columnwidth]{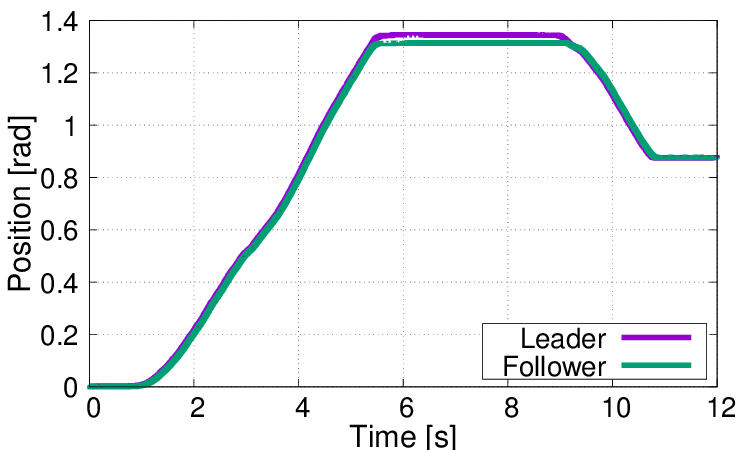}
        \vspace*{-6mm}
        \subcaption{}
        \label{fig:sym_pos}
    \end{minipage}
    \begin{minipage}{0.245\textwidth}
        \centering
        \includegraphics[width=\columnwidth]{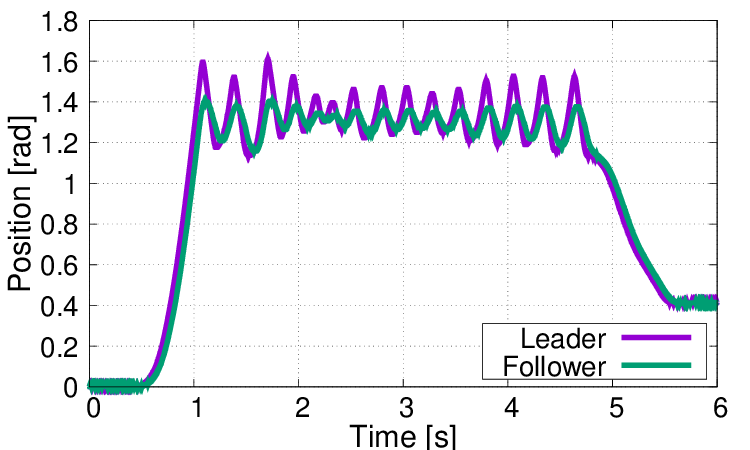}
        \vspace*{-6mm}
        \subcaption{}
        \label{fig:for_pos}
    \end{minipage}
    \begin{minipage}{0.245\textwidth}
        \centering
        \includegraphics[width=\columnwidth]{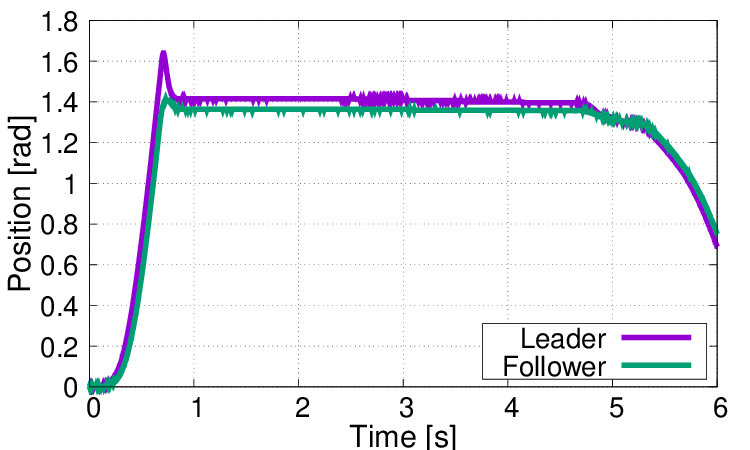}
        \vspace*{-6mm}
        \subcaption{}
        \label{fig:3ch_pos}
    \end{minipage}
    \begin{minipage}{0.245\textwidth}
        \centering
        \includegraphics[width=\columnwidth]{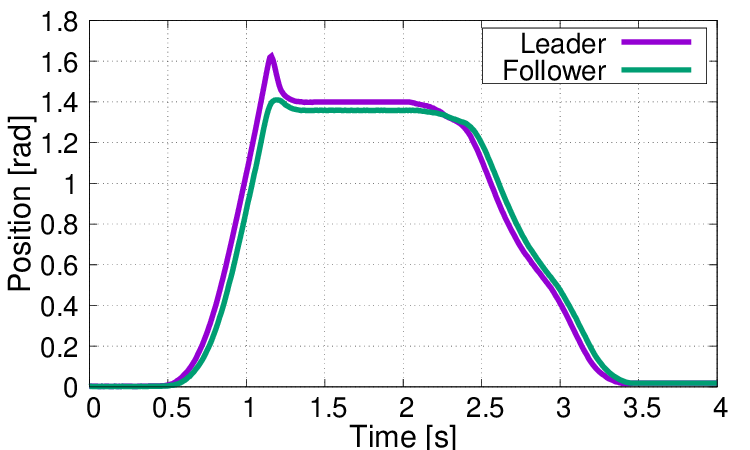}
        \vspace*{-6mm}
        \subcaption{}
        \label{fig:3ch_pos100}
    \end{minipage}\\
    \begin{minipage}{0.245\textwidth}
        \centering
        \includegraphics[width=\columnwidth]{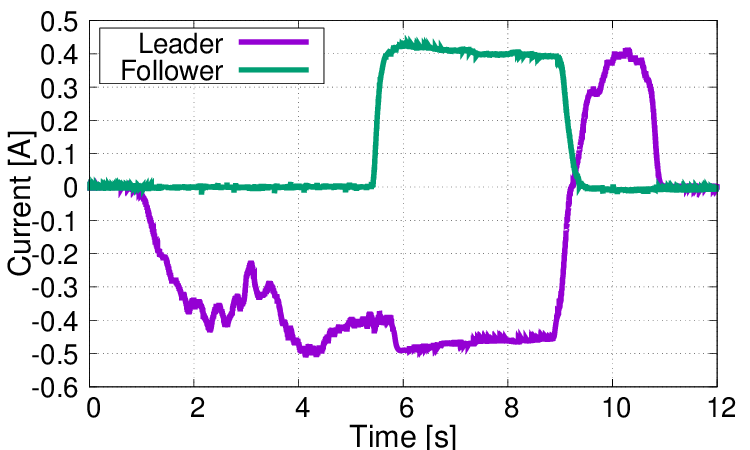}
        \vspace*{-6mm}
        \subcaption{}
        \label{fig:sym_tor}
    \end{minipage}
    \begin{minipage}{0.245\textwidth}
        \centering
        \includegraphics[width=\columnwidth]{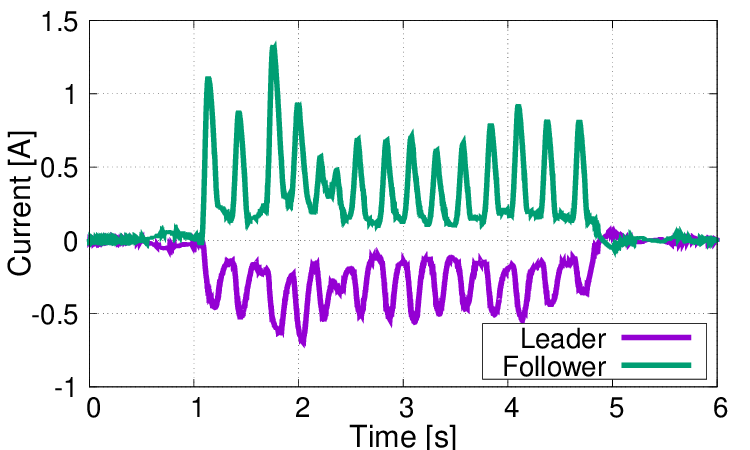}
        \vspace*{-6mm}
        \subcaption{}
        \label{fig:for_tor}
    \end{minipage}
    \begin{minipage}{0.245\textwidth}
        \centering
        \includegraphics[width=\columnwidth]{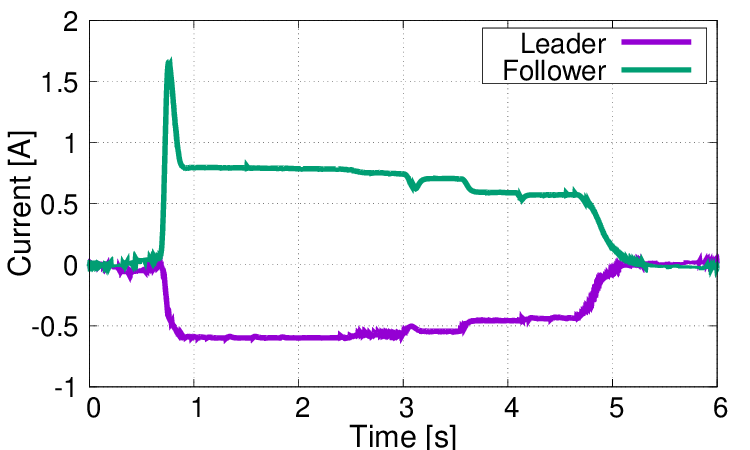}
        \vspace*{-6mm}
        \subcaption{}
        \label{fig:3ch_tor}
    \end{minipage}
    \begin{minipage}{0.245\textwidth}
        \centering
        \includegraphics[width=\columnwidth]{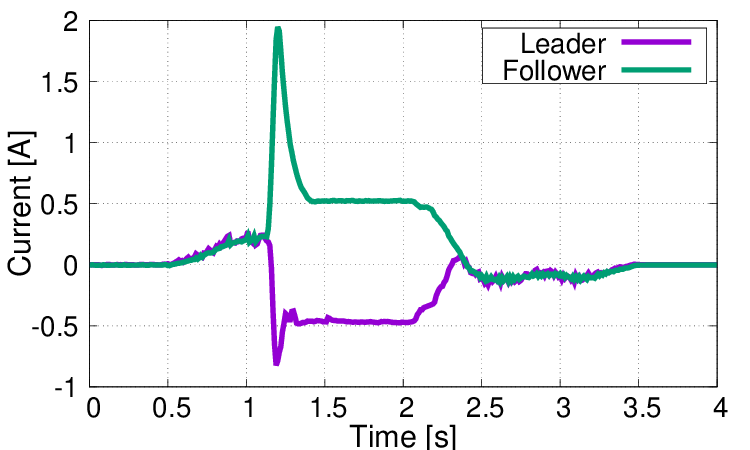}
        \vspace*{-6mm}
        \subcaption{}
        \label{fig:3ch_tor100}
    \end{minipage}

    \caption{Results for Experiments on actuator modules.(a) Positions of SPBT at 1 kHz, (b) Positions of FRBT at 1 kHz, (c) Positions of IGBT at 1 kHz, (d) Positions of IGBT at 100 Hz, (e) Currents of SPBT at 1 kHz, (f) Currents of FRBT at 1 kHz, (g) Currents of IGBT at 1 kHz, (h)Currents of IGBT at 100 Hz.}
    \vspace*{-4mm}
    \label{fig:exp_result}
\end{figure*}

In the experiment, an external force is applied to the leader actuator in the direction indicated by the red arrow in the figure, 
starting from an initial angle of 0 rad. The position and current response values are measured when the follower link contacts 
the obstacle. For comparison, four control schemes are evaluated: (1) SPBT at a high-level system cycle of 1 kHz, 
(2) FRBT at 1 kHz, (3) IGBT at 1 kHz, and (4) IGBT at 100 Hz. 
The leader actuator in FRBT operates in current control mode, while all other schemes use position control mode. 
Control parameters are set to the default values of the Dynamixel actuators, and no parameter tuning is performed.
For the implementation of IGBT, the follower current response value is used instead of the control input $u_f$ 
to dynamically adjust the leader current limitation function. Moreover, since identical actuators are used, 
$\alpha=1$ and $\beta=1$, and gravity effects are negligible, so $u_f^{FF}=0$.

Fig. \ref{fig:exp_result} shows experimental results. 
For SPBT at 1kHz, during the free motion phase ($t<6$), only the leader current reacts, 
indicating that the leader actuator is driven in a direction opposing the applied operational force. 
As a result, it takes approximately 4 seconds for the follower to reach the obstacle, confirming low operability. 
In the contact motion phase ($6<t<9$), the sum of the leader and follower currents is nearly zero, 
demonstrating the establishment of an action-reaction relationship and the realization of force feedback.
For FRBT at 1kHz, during the free motion phase ($t<1$), the leader current remains relatively small, 
and the follower reaches the obstacle in approximately 1 second, indicating improved operability compared to SPBT. 
However, during the contact motion phase ($1<t<5$), both the leader and follower exhibit oscillatory behavior, 
confirming low contact stability.
For IGBT at 1kHz, during the free motion phase ($t<1$), operability is comparable to FRBT. 
Although overshooting is observed at the moment of contact, during the contact motion phase ($1<t<5$), 
the system establishes an action-reaction relationship, demonstrating higher contact stability compared to FRBT.
For IGBT at 100Hz, while the magnitude of overshooting increases at the moment of contact, 
both operability and contact stability remain nearly unchanged.

From the experimental results, it has been demonstrated that IGBT can be implemented on actuator modules designed 
for low-cost hardware by utilizing the follower current response and the leader current limitation function. 
Furthermore, IGBT can be implemented with zero parameter adjustment effort by using the default control parameters as configured. 
The experimental validation also confirmed that IGBT achieves high operability and contact stability simultaneously, 
consistent with numerical simulations, and is robust against variations in the cycle of the high-level system.

\subsection{Experiments on low-cost hardware}
\begin{figure}[t]
    \centering
    \begin{minipage}{0.75\columnwidth}
        \centering
        \includegraphics[width=\columnwidth]{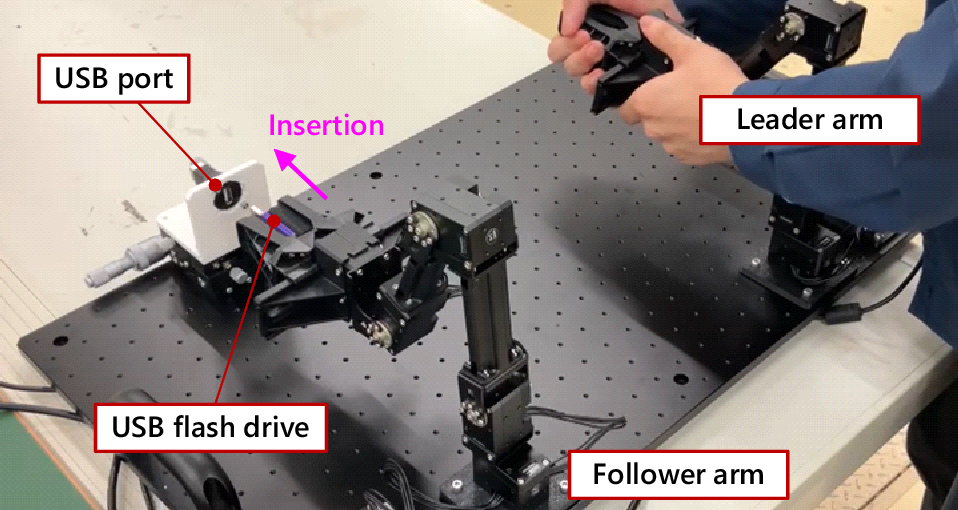}
        \subcaption{}
        \label{fig:om}
    \end{minipage}\\
    \begin{minipage}{0.75\columnwidth}
        \centering
        \includegraphics[width=\columnwidth]{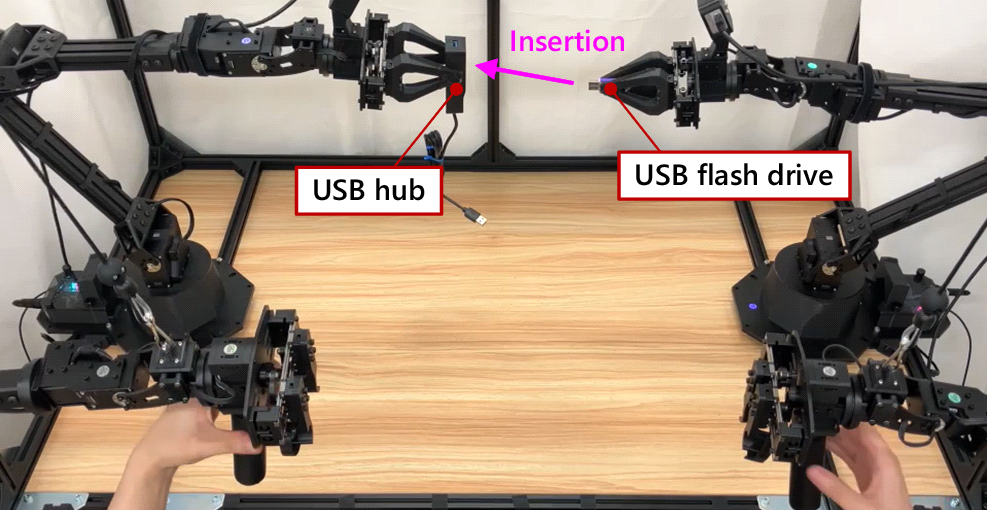}
        \subcaption{}
        \label{fig:aloha}
    \end{minipage}
    \caption{experimental environment for low-cost hardware. (a) OpenMANIPULATOR-X, (b) ALOHA.}
    \vspace*{-4mm}
    \label{fig:hardware_env}
\end{figure}

Finally, we validate the practical effectiveness of IBGT, whose fundamental efficacy and robustness were confirmed in the previous section, 
by implementing it on low-cost multi-joint robots. Compared to the experiments in the previous section, 
this system introduces new challenges such as simultaneous control of multiple motors, the influence of arm self-weight, more complex environmental interaction, 
and reduced rigidity due to inexpensive actuators. Consequently, conventional bilateral teleoperation methods—especially those with feedback controllers placed 
in high-level systems—face increased implementation complexity and parameter tuning difficulty.
As shown in the Fig. \ref{fig:hardware_env}, we set up two types of low-cost hardware, OpenMANIPULATOR-X \cite{om} and ALOHA \cite{ALOHA}. 
To demonstrate IBGT's practical applicability, we performed the complex and contact-rich USB insertion task.

\subsubsection{OpenMANIPULATOR-X}
The bilateral teleoperation between a pair of OpenMANIPULATOR-X robots was evaluated. 
As described in the previous section, IGBT was implemented by utilizing the follower current response 
and the leader current limitation function, with $\alpha=1$ and $\beta=1$. 
To compensate for the effects of the self-weight, $u_f^{FF}$ was sequentially calculated based on the joint angles. 
The relationship between current and torque was determined according to reference \cite{dynamixel}. 
The control parameters were kept at the default values of the Dynamixel actuators, and no parameter tuning was performed.
The cycle rate of the high-level system was set to 100 Hz.
A task was conducted in which the follower gripped a USB flash drive and inserted it into a USB port fixed in the environment.

\begin{figure}[t]
    \centering
    \begin{minipage}{0.492\columnwidth}
        \centering
        \includegraphics[width=\columnwidth]{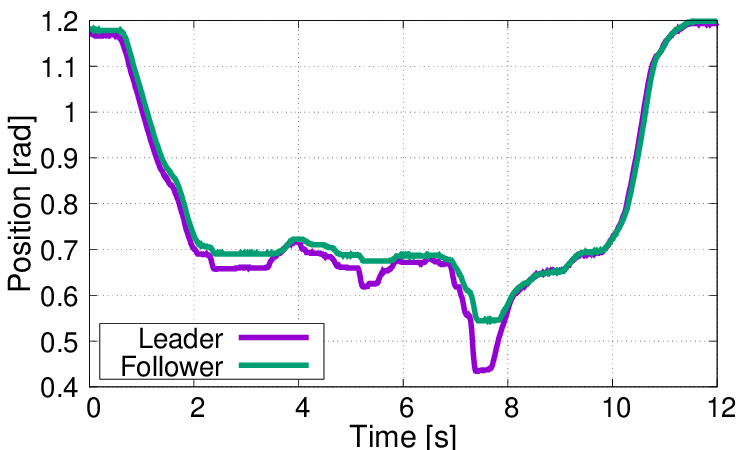}
        \subcaption{}
        \label{fig:ompos}
    \end{minipage}
    \begin{minipage}{0.492\columnwidth}
        \centering
        \includegraphics[width=\columnwidth]{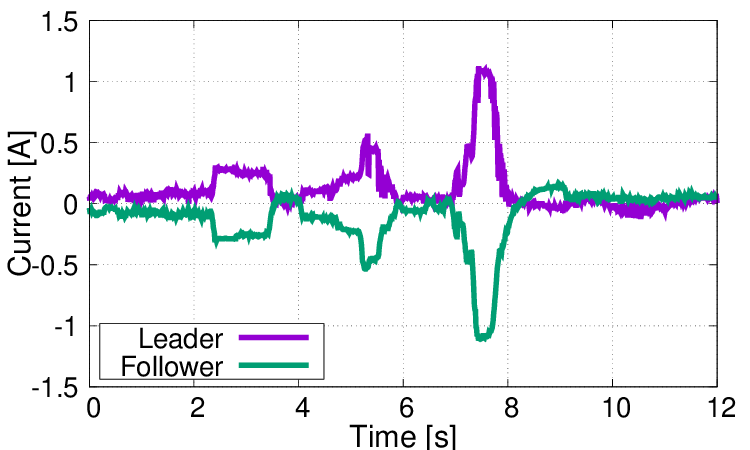}
        \subcaption{}
        \label{fig:omcur}
    \end{minipage}
    \caption{experimental results of the USB Insertion Task Using OpenMANIPULATOR-X. (a) Positions of joint 2, (b) Currents of joint 2.}
    \vspace*{-4mm}
    \label{fig:om_result}
\end{figure}

Fig. \ref{fig:om_result} shows experimental results of the USB Insertion Task Using OpenMANIPULATOR-X.
As a representative, the position and current of Joint 2, where the action-reaction relationship is most apparent, were plotted. 
The current values are those obtained after compensating for the effect of self-weight. During the task execution, 
the USB insertion failed twice and succeeded on the third attempt, which is reflected in the current response. 
In all cases, the action-reaction relationship was consistently established.
Furthermore, we used default control parameter settings without any adjustment, which demonstrates the ease of implementation of IBGT even on multi-joint robots.

\subsubsection{ALOHA}
In the case of ALOHA, the actuators used for the leader and follower differed depending on the joint, 
and thus $\alpha$ was determined based on the ratio of the nominal maximum torque of the respective actuators. 
While the actuator in the leader gripper was not equipped with current measurement or current limitation functionality, 
it was capable of measuring load rates and limiting PWM signals, which were used to determine $\alpha$. 
Additionally, $\beta$ was set for the gripper to adjust the opening width. 
To compensate for the effects of the arm's self-weight, $u_f^{FF}$ was sequentially calculated based on the joint angles.  
The relationship between current and torque was determined according to reference \cite{dynamixel}. 
The control parameters were kept at the default values of the Dynamixel actuators, and no parameter tuning was performed.
The cycle rate of the high-level system was set to 100 Hz.
A task was conducted in which the left follower gripped a USB hub and the right follower gripped a USB flash drive, 
and both were inserted into their respective ports.

\begin{figure}[t]
    \centering
    \begin{minipage}{0.492\columnwidth}
        \centering
        \includegraphics[width=\columnwidth]{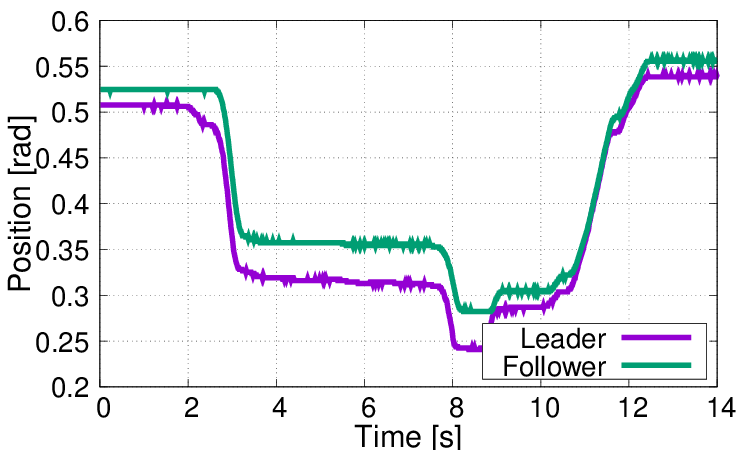}
        \subcaption{}
        \label{fig:alohapos}
    \end{minipage}
    \begin{minipage}{0.492\columnwidth}
        \centering
        \includegraphics[width=\columnwidth]{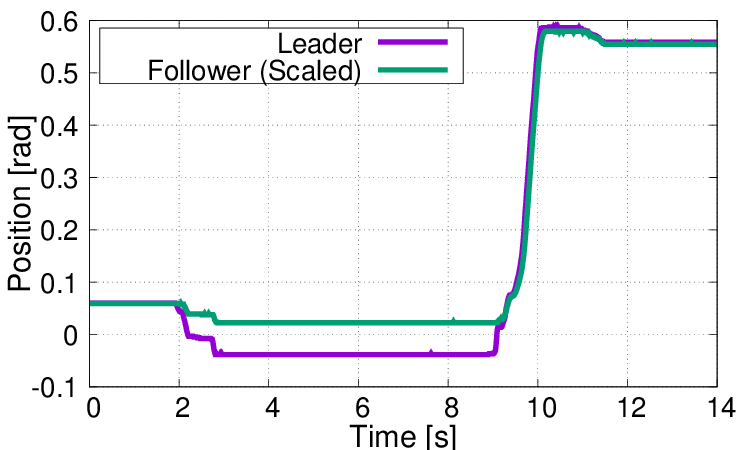}
        \subcaption{}
        \label{fig:aloha_gr_pos}
    \end{minipage}\\
    \begin{minipage}{0.492\columnwidth}
        \centering
        \includegraphics[width=\columnwidth]{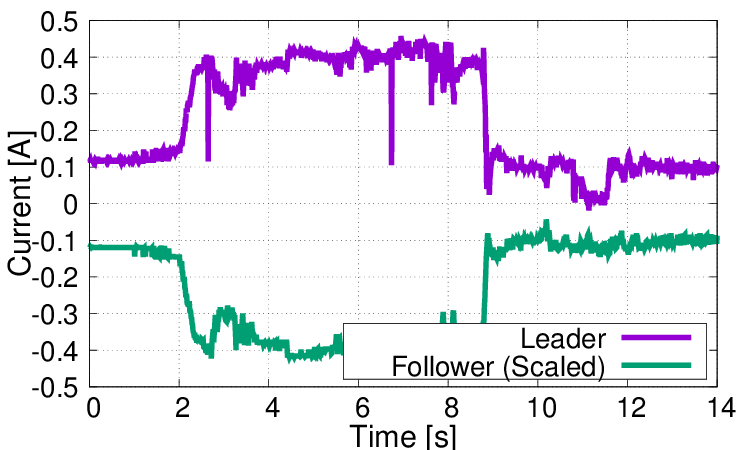}
        \subcaption{}
        \label{fig:alohacur}
    \end{minipage}
    \begin{minipage}{0.492\columnwidth}
        \centering
        \includegraphics[width=\columnwidth]{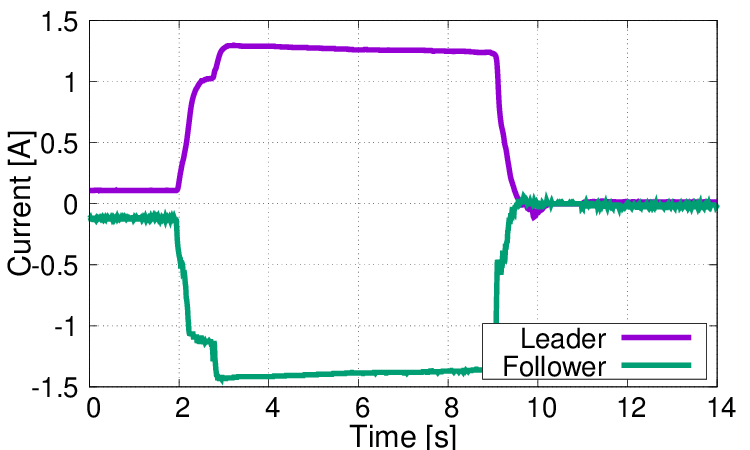}
        \subcaption{}
        \label{fig:aloha_gr_cur}
    \end{minipage}
    \caption{experimental results of the USB Insertion Task Using ALOHA. (a) Positions of left joint 3, (b) Positions of right gripper, (c) Positions of left joint 3, (a) Currents of right gripper.}
    \vspace*{-4mm}
    \label{fig:aloha_result}
\end{figure}

Fig. \ref{fig:aloha_result} shows experimental results of the USB Insertion Task Using ALOHA.
As a representative, the position and scaled current of the left joint 3, 
as well as the scaled position and estimated current of the right gripper, were plotted. 
In all cases, the action-reaction relationship was consistently established, 
demonstrating the effectiveness of IGBT not only between actuators with different output characteristics 
but also across states involving different physical quantities.

From the foregoing, our ability to apply IBGT to two types of low-cost hardware and successfully complete the USB insertion task 
without any parameter adjustment demonstrates the method's high ease of implementation, versatility, and applicability to real-world environments.

\section{Discussion}
In this study, we proposed a bilateral teleoperation method called "IGBT," which can be easily implemented on low-cost hardware, and validated its effectiveness through numerical simulations and experimental evaluations. Below, the significance and limitations of the research findings are discussed.

IGBT is characterized by requiring remarkably low adjustment effort for implementation compared to conventional bilateral teleoperation methods. In particular, leveraging the default settings of position controllers allows it to achieve high performance without parameter tuning, which significantly enhances its practicality for low-cost hardware. Additionally, the configuration of dynamically modifying the leader's current limitation based on the follower's current response demonstrated high adaptability to changes in motion. This simple design has the potential to promote the widespread adoption of force-feedback-enabled teleoperation systems.
Furthermore, experimental results confirmed that IGBT achieves a high level of both operability and contact stability compared to conventional methods such as SPBT and FRBT. Notably, even under slower cycle rates of high-level systems, its performance was scarcely affected, demonstrating robustness—a critical attribute for low-cost hardware applications. Moreover, applicability to different types of low-cost hardware, including OpenMANIPULATOR-X and ALOHA, was verified, showcasing the versatility of IGBT. These findings underscore the practical utility of IGBT for low-cost hardware systems.

However, this study has several limitations. First, IGBT is specifically designed for low-cost hardware, and its effectiveness in high-end robotic systems or systems requiring high-precision force sensors remains unverified. Additionally, it remains unclear whether IGBT offers advantages over methods like ABC, which are rigorously designed with high cycle rates for enhanced operability and contact stability.
Furthermore, this study utilized actuators with high backdrivability, but the applicability of IGBT to hardware with low backdrivability—such as industrial robots whose motor currents are less sensitive to external forces—requires further investigation. Evaluating its suitability for such hardware would potentially expand the scope of IGBT's applicability.

\section{Conclusion}

In this paper, we proposed a bilateral teleoperation method, Input-Gated Bilateral Teleoperation (IGBT), designed for low-cost hardware. IGBT enables easy implementation without the need for force sensors, utilizing a simple low-level controller. Through a series of experiments, we demonstrated that IGBT achieves high operability and contact stability compared to conventional methods, while requiring minimal parameter adjustment effort due to its simple design. Furthermore, IGBT exhibited robustness against variations in the cycle rate of high-level systems.

These findings suggest that IGBT has the potential to significantly simplify the implementation of force-feedback-enabled bilateral teleoperation for low-cost hardware. This, in turn, could expand the applicability of robots to imitation learning and contact-rich tasks, thereby contributing to the broader adoption and advancement of robotic technologies.

\addtolength{\textheight}{-12cm}   



\section*{ACKNOWLEDGMENT}

This work was supported by JST [Moonshot R\&D][Grant Number JPMJMS2031].


\end{document}